\documentclass[10pt,twocolumn,letterpaper]{article}

\usepackage{wacv}
\usepackage{times}
\usepackage{epsfig}
\usepackage{graphicx}
\usepackage{amsmath}
\usepackage{amssymb}
\usepackage{enumitem}

\usepackage{enumitem}
\setlist{nolistsep,leftmargin=*}

\usepackage{graphicx}
\usepackage{comment}
\usepackage{amsmath,amssymb} 
\usepackage{color}

\usepackage[dvipsnames]{xcolor}
\usepackage{latexsym}
\usepackage{lipsum}
\usepackage{amssymb}
\usepackage{amsmath}

\usepackage{epstopdf}
\usepackage{url}
\usepackage{booktabs}
\usepackage{subfig}
\usepackage{subfloat}
\usepackage{graphicx}

\usepackage{array, floatrow, tabularx, makecell, booktabs}

\usepackage[belowskip=-15pt,aboveskip=0pt]{caption}

\newcommand{\SV}[1]{#1}

\newcommand{\GB}[1]{#1}

\newcommand{\SA}[1]{ {#1}}

\def\wacvPaperID{932} 

\wacvfinalcopy 

\ifwacvfinal
\def\assignedStartPage{1} 
\fi

\ifwacvfinal
\usepackage[breaklinks=true,bookmarks=false]{hyperref}
\else
\usepackage[pagebackref=true,breaklinks=true,colorlinks,bookmarks=false]{hyperref}
\fi

\ifwacvfinal
\else
\fi

\begin{document}

\def\our{TVVSD~}
\def\lpc{\emph{lpc}}
\def\lpct{\emph{lpc~}}
\title{Transductive Visual Verb \SA{Sense} Disambiguation} 

\author{Sebastiano Vascon\\
University of Venice\\
{\tt\small sebastiano.vascon@unive.it}
\and
Sinem Aslan\\University of Venice\\                        Ege University
\\
{\tt\small sinem.aslan@unive.it}
\and
Gianluca Bigaglia\\
University of Venice\\
{\tt\small gianluca.bigaglia@unive.it}
\and
Lorenzo Giudice\\
University of Venice\\
{\tt\small lorenzo.giudice@unive.it}
\and
Marcello Pelillo\\
University of Venice\\
{\tt\small marcello.pelillo@unive.it}
}

\maketitle

\begin{abstract}
Verb Sense Disambiguation is a well-known task in NLP, the aim is to find the correct sense of a verb in a sentence. Recently, this problem has been extended in a multimodal scenario, by exploiting both textual and visual features of ambiguous verbs leading to a new problem, the Visual Verb Sense Disambiguation (VVSD). Here,
the sense of a verb is assigned considering the content of an image paired with it rather than a sentence in which the verb appears. Annotating a dataset for this task is more complex than textual disambiguation, because assigning the correct sense to a pair of $<$image, verb$>$ requires both non-trivial linguistic and visual skills. In this work, differently from the literature, the VVSD task will be performed in a transductive semi-supervised learning (SSL) setting, in which only a small amount of labeled information is required, reducing tremendously the need for annotated data. The disambiguation process is based on a graph-based label  propagation method which takes into account mono or multimodal representations for $<$image, verb$>$ pairs. Experiments have been carried out on the recently published dataset VerSe, the only available dataset for this task. The achieved results outperform the current state-of-the-art by a large margin while using only a small fraction of labeled samples per sense\footnote{Code available: \url{https://github.com/GiBg1aN/TVVSD}}.
\end{abstract}

\vspace{-0.2cm}
\section{Introduction}

Every language has ambiguous words, e.g., in English, the word \emph{apple} can be referred to as either an IT company, a fruit, or a city. 
Word Sense Disambiguation (WSD) is a common task in natural language processing \cite{navigli2009word}, where the goal is to automatically recognize the correct sense of a word within a sentence. 

Verb Sense Disambiguation (VSD) is a sub-problem of WSD where the correct sense of a \textit{verb} in a sentence is aimed to be identified \cite{sudarikov2016verb}.
For instance, while 
the most common sense of the verb \textit{run} is the one related to moving quickly,
it might have a different sense regarding to its context, such as the one 
related to machine operations (\textit{the washing machine is running}) or covering a distance (\textit{this train runs hundreds of miles every day}); all these senses share the same verb, but they have quite different meanings.

VSD is an utmost important task, affecting different domains. For example, in an NLP retrieval scenario, it is required the search engine to group 
the results by senses, hence disambiguate the verb senses in queries to retrieve the correct results \cite{di2013clustering}. VSD also takes an important role in other NLP tasks such as, machine translation \cite{sudarikov2016verb}, semantic role labeling \cite{alfano2019neural} and question answering \cite{novischi2006question}. 

In addition to the typical NLP tasks, VSD can be brought to a Computer Vision (CV) domain, taking into account problems like Action Recognition (AR) and Human Object Interaction (HOI) \cite{7912210,8354149}; 
the authors exploit the identification of objects and entities in an image to infer either the action that is being performed or the correct verb that links those entities and objects. Even if there are some clear overlappings between VSD and AR/HOI, the latter do not take into account the ambiguity of verbs. 
The analogy between these tasks in NLP and CV fields can be exploited by combining the features of both domains to improve a disambiguation system's overall performances. Motivated by this fact, recently, \cite{gella2019disambiguating} introduced the \emph{Visual Verb Sense Disambiguation} (VVSD). 

In a VVSD task, the goal is to disambiguate the sense of a verb paired with an image. 
Differently from a standard NLP disambiguation task, in which the context is provided by a phrase, here the context is provided by an image. 

In \cite{gella2019disambiguating} the authors proposed the first (and only) well curated dataset to assess algorithms' performances for VVSD tasks. The reported baselines comprise both supervised and unsupervised models using both unimodal (textual or visual) and multimodal features (textual and visual).

Annotating a dataset for this task is very expensive, since it requires both non-trivial language and visual knowledge \cite {pasini2020train}.
Toward this direction, in this work, we tackle the multimodal VVSD problem, offering a new perspective based on semi-supervised learning which brings significant performance gain at a lower labeling-cost.
The strength of SSL 
algorithms arises when the available labeled set size is not significant to train a fully-supervised classifier or when annotating a full dataset is too expensive or unfeasible. In SSL, only a small amount of labeled data is needed because both labeled and the unlabeled samples embeddings are exploited during inference. 
Thus, we assume to have a small amount of labeled data ($<$image,verb$>$ and its sense) to infer the senses of the unlabeled ones. 

Among the possible SSL algorithms \cite{Zhu02learningfrom}, we choose a game-theoretic model called \emph{Graph Transduction Games} (GTG) \cite{erdem2012graph}. The GTG has been succesfully applied in many different SSL contexts, like deep metric learning \cite{elezi_ECCV}, matrix factorization \cite{tripodi2016context}, image recognition \cite{aslan2020}, protein-function prediction \cite{VASCON2018} and, indeed, traditional text-based WSD setting \cite{tripodi2017game}. Moreover, it works consistently better \cite{vascon2019unsupervised} than other graph-based SSL methods like Label Propagation \cite{Zhu02learningfrom} and Gaussian Fields \cite{zhu2003semi}.

\SV{Our contributions are thus three-fold:
\begin{enumerate}
    \item We proposed a new model for multimodal visual verb sense disambiguation based on semisupervised learning.
    \item We reported an extensive ablation study on the effect of using an increasing number of labeled data.
    \item We outperformed the state-of-the-art by a large margin exploiting only a small fraction of labeled data. 
\end{enumerate}}

\begin{figure*}[ht!]
\footnotesize
\begin{center}
\includegraphics[width=0.8\textwidth]{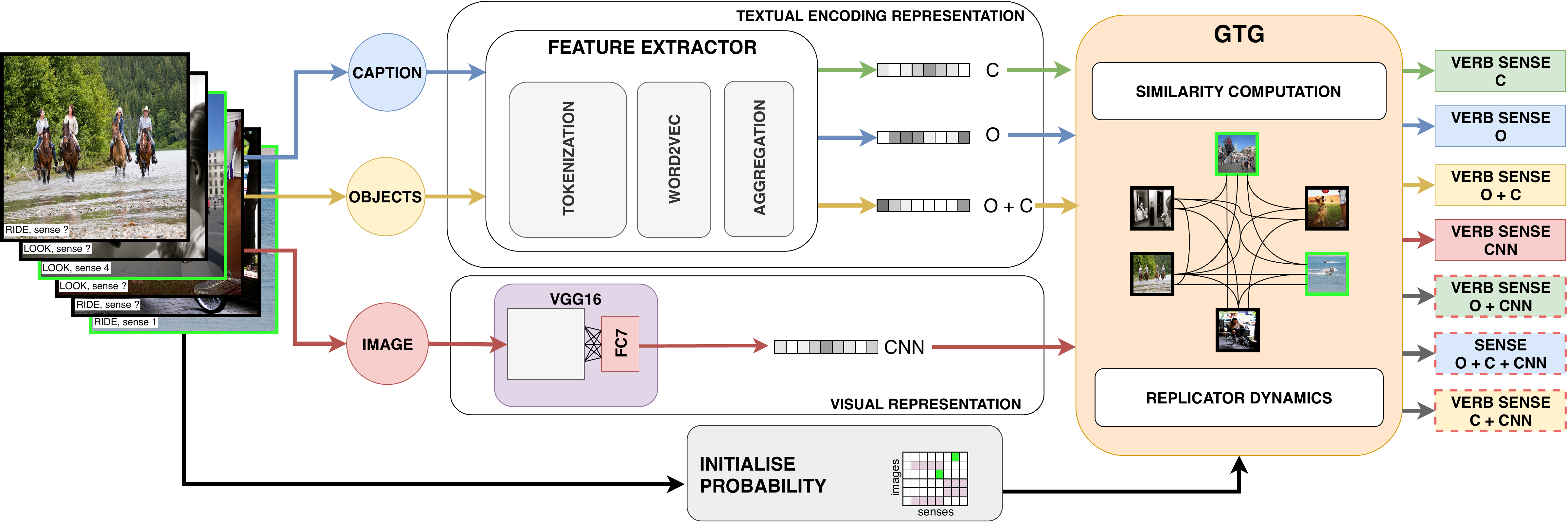}
\caption{\footnotesize Pipeline of the algorithm considering both labeled (green border) and unlabeled images (black border). The letters $O$ and $C$ stands for features generated from Objects label and Caption.}
\label{fig:pipeline}
\end{center}
\vspace{-0.2cm}
\end{figure*}

\vspace{-0.2cm}
\section{Related works}
\vspace{-0.2cm}
\SA{Common approaches for WSD/VSD can be grouped into three categories: \textit{supervised}, \textit{unsupervised} and \textit{knowledge-based} methods. Supervised methods \cite{melamud2016context2vec} rely on sense-tagged corpora which act as the training set.} Such algorithms usually exploit linguistic features like: $n$-grams that surround a word, syntactical dependencies tags (e.g. subject, object, verb) or context information summarized in structures like co-occurrence matrices. 
Performance of supervised methods is hindered by the requirement of handling all the possible senses of target corpus while it is implausible to have a training set with a sufficiently big sample size for each word/verb sense \cite{navigli2009word}. Thus, unsupervised learning algorithms, that do not exploit any training data, may be a more suitable solution when the number of senses to handle becomes unfeasible. Purely unsupervised methods rely on the distribution of senses and exploit the fact that words/verbs with the same sense would have similar contextual information. However, while they extract clusters of senses, they do not rely on exact sense labeling of words/verbs which would yield to extracted senses are likely to not match the ones categorized and defined in standard dictionaries. On the other hand at the knowledge-based methods, rather than extracting the sense inventory from the corpus, it is known a-priori. So, a mapping between a dictionary and the occurrences in the corpus is performed and by relying on lexical databases \cite{pradhan2009ontonotes,miller1990introduction,navigli2010babelnet} semantic accordance is used to disambiguate. In \cite{tripodi2017game}, a semi-supervised learning model for WSD is proposed, facing a pure textual, and not multimodal, task.

While there is a huge literature on Word Sense Disambigutation (WSD) adopting (unimodal) textual features, visual clues for WSD in a multimodal setting was studied by limited works. One of the first approaches is in \cite{barnard2005word} which used a statistical model based on joint probabilities between images and words. Following an unsupervised approach, \cite{loeff2006discriminating} applied spectral clustering for image sense disambiguation; while \cite{chen2015microsoft} applied co-clustering through textual and visual domain to explore multiple senses of a given noun phrase.
\cite{saenko2009unsupervised} used LDA to discover a latent space by exploiting dictionaries definitions to learn distributions that represent senses. A similar task was accomplished in \cite{berzak2016you} that tried to solve linguistic ambiguities using multimodal data. In \cite{botschen2018multimodal} they used multimodal data for semantic frame labeling. Performances of all these aforementioned works are quite good, however such techniques are noun-oriented and perform poorly for verb disambiguation tasks. The first attempt to perform a fully verb-oriented sense disambiguation 
was introduced in \cite{gella2016unsupervised}, which designed a variation of Lesk algorithm \cite{lesk1986automatic} that uses the multimodal sense encoding and the multimodal input encoding respectively as the definition and context for the algorithm. 

\vspace{-0.2cm}
\section{Transductive VVSD}
\vspace{-0.2cm}

In this section we dissect the different components of our model, named \textit{Transductive Visual Verb Sense Disambiguation} (\our). The global picture can be seen in Fig \ref{fig:pipeline}.\\ 
Our model is made of four steps:
\begin{enumerate}
\item Feature extraction for each pair $<$image, verb$>$.
\item Construction of a graph-based representation of all pair $<$image, verb$>$.
\item Initialization of the assignment between $<$image, verb$>$ and possible senses.
\item Transductive inference via dynamical system assigning $<$image, verb$>$ to a sense.
\end{enumerate}
in the following we will consider the $i$-th pair $<$image, verb$>$ as an atomic entity.

\subsection{Feature extraction}\label{sec:features}
We follow the schema proposed in \cite{gella2019disambiguating} for a fair comparison with the state-of-the-art, although more recent feature models can be easily plugged in our pipeline. For each pair $<$image, verb$>$ we extract the following embeddings:\\
\textbf{Visual features.} Input images are fed into a pre-trained VGG16\footnote{We used the PyTorch implementation of VGG} model \cite{simonyan2014very}, and the output of the last FC layer is used as feature representation, resulting in a vector of 4096 elements for each image. Such vector is then unit-normalized.\\
\textbf{Textual features.} \GB{As in \cite{gella2019disambiguating}, experiments on text have been run on two possible setups: using VerSe textual data \SV{annotations} (GOLD) or by generating them through state-of-art object detectors and image descriptors (PRED). In the latter scenario, object labels have been predicted using a VGG16 model. \SV{Since the VGG16 net classifies images without performing object detection, in \cite{gella2019disambiguating} they thresholded the output of the SoftMax layer taking only \SV{classes} that had a score greater than 0.2 (or the highest in case of empty result). This allows to obtain multiple classes/objects per image.} 

Captions have been generated with NeuralTalk2\footnote{\url{https://github.com/karpathy/neuraltalk2} \cite{karpathy2015deep}} \cite{vinyals2015show}.} For what concerns the encoding, either captions, objects labels or a concatenation of both can be used. 
The encoding is performed through word2vec \cite{mikolov2013efficient} embedding in a 300-dimensional space. It is based on a Gensim model \cite{rehurek_lrec} pre-trained on Google News dataset. For each word composing the textual input, a word2vec embedding is extracted. After that, they are aggregated by mean and unit-normalized, resulting in a vector for each image.\\

\textbf{Multimodal features.} To perform multimodal VSD, textual and visual features are combined. In \cite{gella2019disambiguating}, beyond the vector concatenation, Canonical Correlation Analysis and Deep Canonical Correlation Analysis are also explored. Nevertheless, their performances were poorer than concatenation ones, \SV{hence we explored only this last option}.

\subsection{Graph Construction}
The core of our method relies on a graph-based semi-supervised learning algorithm, named \emph{Graph Transduction Games} (GTG) \cite{erdem2012graph}. Such method requires as input a weighted graph $G$, in which a set of labeled $L$ and unlabeled nodes $U$ are present, and a stochastic initial assignment $X$ between nodes to labels (senses). The output is then a refined assignment matrix $X$ which is the results of nodes interaction. 

After extracting the desired embedding (visual, textual or multimodal), we construct a weighted graph $G=(V,E,\omega)$ with no self-loop over all the items in the dataset.
Here $V$ corresponds to all the pair $<$image, verb$>$ in both set $L$ and $U$, hence $V=L \cup U$. The set of edges $E \subseteq V \times V$ connects all the nodes and the function $\omega: e \in E \rightarrow \mathbb{R}_{\geq 0}$ weighs the pairwise similarity between vertices.

We define the similarity $\omega$ between node $i$ and $j$ (the edges weight), as the cosine\footnote{Since the features are all non-negative with unit norm, the cosine can be computed using the dot product.} of their $d$-dimensional features embedding $f_i$ and $f_j$:
$$
\footnotesize
\omega_{i,j} = \begin{cases}
        \sum_{i=1}^{d}{f_{i,d}\cdot f_{j,d}} \quad \mbox{ if $i \neq j$} \\
        0 \quad \mbox{ otherwise}
        \end{cases}
$$

 Within this context, $f_i$ is computed considering one of the modalities presented above. In the experimental section we report performances for all the embeddings (mono-modality) and their combinations (multi-modality).
 All the pairwise similarities $\omega_{i,j}$ are stored in a matrix $W \in \mathbb{R}^{n \times n}$.

\subsection{Initial assignment}
The goal of the transductive process is to propagate the labels from the labeled set $L$ to the unlabeled ones $U$. For this purpose each node $i \in V$ is paired with a probability vector $x_i$ over the possible senses ($x_i \in \Delta^m$ where $m$ is the number of senses and $\Delta^m$ is standard $m$-dimensional simplex). Such vector is initialized in two different ways, based on fact that it belongs to a labeled or an unlabeled node.
For the labeled node:
     $$
        \footnotesize
        x_{i,h}=
        \begin{cases}
        1 \quad \mbox{ if $i$  have sense $h$} \\
        0 \quad \mbox{ otherwise}
        \end{cases}
    $$ 
while for the unlabeled nodes:
     $$
        \footnotesize
        x_{i,h}=
        \begin{cases}
        \frac{1}{|S_i|} \quad \mbox{ if $h \in S_i$} \\
        0 \quad \mbox{ otherwise}
        \end{cases}
    $$ 
where $x_{i,h}$ corresponds to the probability that the $i$-th node chooses the label $h$, while $S_i$ is the set of possible senses associated with the verb \SV{in the $i$-th node}. All the assignment $x_i$ with $i=\{1 \dots n\}$ are stored into a stochastic matrix $X \in \mathbb{R}^{n \times m}$.

\subsection{Transductive Inference}
\vspace{-0.2cm}
The transductive inference is performed with a dynamical system, which is responsible to iteratively refine the initial assignment $X$. We define here two quantities:
\vspace{-0.3cm}
\begin{eqnarray}
u_{i,h} = \sum_{j \in U}{(A_{ij} x_j)_h} + \sum_{k = 1}^m{\sum_{j \in L_k}{A_{ij}(h, k)}} \\
u_i = \sum_{j \in U}{x_i^{T}A_{ij} x_j}+\sum_{k = 1}^m{\sum_{j \in L_k}{x_i^{T} (A_{ij})_k}}
\vspace{-0.3cm}
\end{eqnarray}
where $L_k$ is the set of nodes labeled with class $k$. The matrix $A_{ij} \in \mathbb{R}^{m \times m}$, is defined as $A_{ij} = I_m \times \omega_{ij}$ with $I_m$ being the identity matrix and $\omega_{ij}$ the similarity of nodes $i$ and $j$. The equation $u_{i,h}$ quantifies the support provided by the other nodes to the labeling hypothesis $h$ for the node $i$. While the equation $u_i$ quantifies the overall support received to the node $i$ by the other nodes.

In the following, we add the time component $t$ to distinguish between different iterative steps. For example, $x_i^{(t)}$ refers to the probability vector $x_i$ at time $t$. The dynamical system, responsible for the assignment refinement, is formulated as follow:
\vspace{-0.3cm}
\begin{equation}
    x_{i,h}^{(t + 1)} = x_{i,h}^{(t)}\frac{u_{i,h}^{(t)}}{u_i^{(t)}}
    \label{eq:RD}
    \vspace{-0.3cm}
\end{equation}
The Eq. \ref{eq:RD} is repeated until all the vectors $x_i$ stabilize. Such dynamical system is known as \emph{replicator dynamics} \cite{smith1982evolution} and mimics a natural selection process in which better-than-average hypothesis get promoted while others get extinct. It is worth noting that the refinement takes into account all the hypotheses of all the nodes. In this sense, the labeling is performed not in isolation but is the result of nodes interactions. The rationale is that similar nodes tend to have the same label. The more two nodes are similar, the more they will affect each other in picking the same class.

The Eq. \ref{eq:RD} grants that the matrix $X$ at convergence is a \emph{labeling consistent} solution \cite{hummel1983foundations}\cite{pelillo1997dynamics}. A weighted labeling assignment $X$ is said to be consistent if:
\vspace{-0.3cm}
$$
\sum_{h=1}^m{x_{i,h}u_{i,h}} \geq \sum_{h=1}^m{y_{i,h}u_{i,h}} \mbox{  } \forall i=1, \dots, n
$$
for all $Y$. This means that no other solution $Y$ can perform better than $X$.

Finally, it is worth noted that Eq. \ref{eq:RD} can be written in a matricial form for a fast GPU implementation:
\begin{equation}
    x_i(t + 1) = \frac{x_i(t) \odot (W x(t))_i}{x_i(t)(W x(t))_i^T}
\end{equation}
where $\odot$ represents the Hadamard (element-wise) product.

Regarding the Eq.\ref{eq:RD}, 10 iterations are typically sufficient to reach convergence \cite{ELEZI2018}.

\vspace{-0.2cm}
\section{Experiments}
\vspace{-0.2cm}
In this section, we reported the performances and the experimental settings of our proposed model, TVVSD. The experiments have been carried out on the only available benchmarks for this task, the VerSe and VerSe-19verbs datasets, following the same evaluation protocol as \cite{gella2019disambiguating}.
\subsection{Datasets}
\vspace{-0.2cm}
The VerSe dataset  \cite{gella2019disambiguating} is composed of images selected from Common Objects in Context (COCO) \cite{chen2015microsoft} and Trento Universal Human Object Interaction (TUHOI) \cite{le2014tuhoi}, 90 verbs and 163 possible senses, resulting in 3510 (image, verb) pairs. Verbs have been categorized as \textit{motion} and \textit{non-motion} based on Levin verb classes \cite{levin1993english}, \SV{resulting} in 39 motion and 51 non-motion verbs.

Further, we reported performances on a subset of VerSe, named \emph{VerSe-19verbs}, which is composed of verbs that have at least 20 images and at least two senses in the VerSe dataset, resulting in 19 motion and 19 non-motion verbs. 
\subsection{Competitors and baselines}
\vspace{-0.2cm}
We compare our method with two state-of-the-art algorithms: Gella et al. \cite{gella2019disambiguating} and Silberer et al. \cite{silberer2018grounding}. To the best of our knowledge \cite{gella2019disambiguating} and \cite{silberer2018grounding} are the only literature works dealing with the research problem of VVSD and reported performances on the VerSe dataset. 
The work of \cite{gella2019disambiguating} is based on a variant of Lesk algorithm \cite{lesk1986automatic} in which the sense is assigned based on the cosine similarity between the embedding of $<$image, verb$>$ and the possible verb-senses in the dictionary. This procedure does not require labeled data since the final choice is based on the maximum score between $<$image, verb $>$ and all the possible senses associated to the same verb as the image. For this reason, we tagged this method as \emph{unsupervised}.

\cite{gella2019disambiguating} proposed a supervised setting, in which a logistic regression is trained on each different embeddings. Similarly to \cite{gella2019disambiguating}, in \cite{silberer2018grounding} a logistic regression is trained, but it uses a frame-semantic image representation of the $<$image, verb$>$ pairs rather than the embedding generated by \cite{gella2019disambiguating}. Both methods are categorized as \emph{supervised} and performances are reported only on the VerSe-19verb dataset.

\begin{table*}[ht!]
\vspace{-1cm}
\centering
\resizebox{\columnwidth}{!}{
\begin{tabular}{llllllllllll}
\hline
\multicolumn{11}{c}{Using GOLD annotations for objects and captions } \\
\hline
&  &  &  & \multicolumn{3}{c}{Textual} & Visual & \multicolumn{3}{c}{Concat (CNN+)} \\ 
\cline{5-7} \cline{9-11}
& Images & FS$^{*}$ & MFS$^{*}$ & O & C & O+C & CNN & O & C  & O+C  \\ 
\hline
Motion - Unsupervised Gella et al. \cite{gella2019disambiguating} & 1812 & 70.8 & 86.2 &  54.6 & 73.3 & 75.6 & 58.3 & 66.6 & 74.7 & 73.8 \\
Motion - \our (1 lab/sense) & 1812 & 70.8 & 86.2 & \textbf{73.3$\pm$4.4} & {73.4$\pm$6.1} & 74.2$\pm$5.5 & \textbf{73.3$\pm$5.9} & \textbf{74.7$\pm$3.4} & 74.6$\pm$5.3& 74.1 $\pm$ 5.5\\
Motion - \our (2 lab/sense) & 1812 & 70.8 & 86.2 & \textbf{79.4$\pm$4.4} & \textbf{83.1$\pm$2.7} & \textbf{83.3$\pm$2.6}& 7\textbf{8.8$\pm$4.6}& \textbf{80.7$\pm$4.5} & \textbf{84.0$\pm$2.9}& \textbf{83.6$\pm$3.2}\\
Motion - \our (20 lab/sense) & 1812 & 70.8 & 86.2 & \textbf{97.1$\pm$0.8} & \textbf{92.8$\pm$0.06} & \textbf{92.8$\pm$0.06} & \textbf{95.9$\pm$0.06} & \textbf{96.8$\pm$1.0} & \textbf{92.8$\pm$0.05} & \textbf{92.9$\pm$0.1} \\
\hline
NonMotion - Gella et al. \cite{gella2019disambiguating} & 1698 & 80.6 & 90.7 & 57.0 & 72.7 & 72.6 & 56.1 & 66.0 & 72.2 & 71.3 \\
NonMotion - \our (1 lab/sense) & 1698 & 80.6 & 90.7 & \textbf{71.7$\pm$4.7} & 71.3$\pm$4.4 & 75.8$\pm$4.0& \textbf{64.4$\pm$6.6} & \textbf{71.4$\pm$4.9} & 70.8 $\pm$4.7&74.8$\pm$4.0 \\
NonMotion - \our (2 lab/sense) & 1698 & 80.6 & 90.7 & \textbf{82.5$\pm$3.2} & \textbf{81.8$\pm$2.9} & \textbf{82.2$\pm$4.0} & \textbf{80.8$\pm$4.0} & \textbf{83.4$\pm$2.8} & \textbf{81.8$\pm$2.0} & \textbf{81.7$\pm$3.0}\\
NonMotion - \our (20 lab/sense) & 1812 & 80.6 & 90.7 & \textbf{92.0 $\pm$0.8} & \textbf{91.8$\pm$0.7} & \textbf{91.6$\pm$0.4} & \textbf{94.0 $\pm$2.2} & \textbf{92.1$\pm$1.3} & \textbf{92.0$\pm$1.1} & \textbf{91.9$\pm$1.0} \\
\hline
\hline
\multicolumn{11}{c}{Using PRED annotations for objects and captions } \\
\hline
&  &  &  & \multicolumn{3}{c}{Textual} & Visual & \multicolumn{3}{c}{Concat (CNN+)} \\ 
\cline{5-7} \cline{9-11}
& Images & FS$^{*}$ & MFS$^{*}$ & O & C & O+C & CNN & O & C  & O+C  \\ 
\hline
Motion - Unsupervised Gella et al. \cite{gella2019disambiguating} & 1812 & 70.8 & 86.2 & 65.1 & 54.9 & 61.6 & 58.3 & 72.6 & 63.6 & 66.5 \\
Motion - \our (1 lab/sense) & 1812 & 70.8 & 86.2 & \textbf{71.2$\pm$6.4} & \textbf{71.5$\pm$3.9} & \textbf{71.9$\pm$5.1} & \textbf{73.3$\pm$5.9} & 73.0$\pm$6.3 & \textbf{73.8$\pm$4.3}& \textbf{74.0$\pm$3.6} \\
Motion - \our (2 lab/sense) & 1812 & 70.8 & 86.2 & \textbf{79.5$\pm$4.9} & \textbf{77.1$\pm$3.9}& \textbf{80.2$\pm$4.4}& \textbf{78.7$\pm$4.6}  & \textbf{80.2$\pm$4.4} & \textbf{77.7$\pm$3.6}& \textbf{78.3$\pm$3.7} \\
Motion - \our (20 lab/sense) & 1812 & 70.8 & 86.2 & \textbf{94.4$\pm$0.4} & \textbf{92.7$\pm$0.1}& \textbf{92.8$\pm$0.2} & \textbf{95.9$\pm$0.6}& \textbf{94.1$\pm$0.5} & \textbf{92.9$\pm$0.2}& \textbf{92.9$\pm$0.3}\\
\hline
NonMotion - Gella et al. \cite{gella2019disambiguating} & 1698 & 80.6 & 90.7 & 59.0 & 64.3 & 64.0 & 56.1 & 63.8 & 66.3 & 66.1 \\
NonMotion - \our (1 lab/sense) & 1698 & 80.6 & 90.7 & \textbf{64.4$\pm$5.2}& \textbf{72.2$\pm$5.4} & \textbf{73.3$\pm$4.4}& \textbf{64.4$\pm$6.6} & 65.6$\pm$6.0& \textbf{73.3$\pm$4.9} & \textbf{73.3$\pm$4.6}\\
NonMotion - \our (2 lab/sense) & 1698 & 80.6 & 90.7 &\textbf{75.3$\pm$4.1} & \textbf{84.3$\pm$3.3} & \textbf{77.3$\pm$3.4} & \textbf{80.8$\pm$4.0} &\textbf{77.3$\pm$3.4} & \textbf{84.3$\pm$3.7}& \textbf{83.1$\pm$3.7}\\
NonMotion - \our (20 lab/sense) & 1698 & 80.6 & 90.7 & \textbf{93.2$\pm$1.4}& \textbf{92.8$\pm$2.0}& \textbf{92.4$\pm$1.7}& \textbf{95.9$\pm$0.6}& \textbf{93.0$\pm$1.7}& \textbf{92.7$\pm$2.0}& \textbf{92.6$\pm$2.0}\\
\hline
\end{tabular}}
\caption{Accuracy scores on VerSe dataset using different sense and image representations. The bolds are relative to the performances of \cite{gella2019disambiguating}. $^{*}$ FS and MFS can be considered as unsupervised and supervised references respectively.} \label{tab:VerSe}
\vspace{-0.2cm}
\end{table*}


\SA{The performances of the first sense (FS) and most frequent sense (MFS) heuristics are shown in table \ref{tab:VerSe} and \ref{tab:verb19}. Both are widely used in the NLP literature \cite{navigli2009word} and considered respectively as baselines for unsupervised and supervised scenarios in \cite{gella2019disambiguating}. The FS corresponds to the first sense of a verb in the dictionary representing the common sense in a language (not in a specific dataset), while MFS is the most frequent sense in a dataset. MFS is considered as a supervised heuristic since all labeled data are needed to compute it.}
\subsection{Evaluation protocol and metric}
\vspace{-0.2cm}
We used the \SA{same evaluation metric \SV{as the competitors}, that is the predicted sense accuracy. \SV{Being our setting semi-supervised, the accuracy is computed only on the unlabeled part leaving aside the labeled set.}
\SV{The accuracy is } assessed \SV{considering} two forms of textual annotations \cite{gella2019disambiguating} for object labels and descriptions: \textit{GOLD} and \textit{PRED} (see Sec. \ref{sec:features} - Textual features).} 

\vspace{-0.2cm}
\subsection{Textual and visual representation features.}
\vspace{-0.2cm}
\SV{Considering the different features (see Sec \ref{sec:features}) and their combinations there are 7 possible setups for the experiments, which are in line with \cite{gella2019disambiguating}: captions (C), object labels (O), captions with object labels (C+O), CNN features (CNN), CNN features concatenated to captions (CNN+C), CNN features concatenated to object labels (CNN+O) and CNN features concatenated to captions with object labels (CNN+O+C).} 
\vspace{-0.2cm}
\subsection{Experimental setting}
\vspace{-0.2cm}
Here we describe our experimental setting to make the reported results reproducible. Being our model semi-supervised, we carried out experiments considering an increasing number of labeled samples, from 1 up to 20.

The labeled set is generated by random sampling the dataset. To adhere with the evaluation protocol used by our competitors, the sampling is performed differently whether the dataset used is VerSe or VerSe-19verbs.\\
\textbf{VerSe setup:} For VerSe, we perform sampling per \emph{sense}. 
\SA{Since we do not need to tune any parameter in our method, the remaining samples constitute the unlabeled set where we compute the accuracy scores. For the experiments on the VerSe dataset, we sampled up to $n=20$ elements per class since our performances were converging afterward.}\\
\textbf{VerSe-19verbs setup:} For this dataset, the sampling is performed per \emph{verb} \cite{gella2019disambiguating}. To adhere to the competitors \cite{gella2019disambiguating,silberer2018grounding} we used the same split ratios $80/10/10$\footnote{The authors of  \cite{gella2019disambiguating} sent us these quantities since they were not specified in the paper.} for train/val/test but\SV{, since we don't have a real training phase, our labeled set is composed by the 80\% of image-verb and the remaining part (20\%) becomes the unlabeled set.} \SA{We experimented up to sampling $80\%$ of the minimal verb class which contains 20 images, thus experimented up to $n=16$ labeled samples per verb class.} 
The points which are not sampled are considered as unlabeled and the final accuracy is computed accordingly.

To account for data variability in the sampling process, we performed the experiments 15 times using different random-seeds and reported means and standard deviation.

\section{Results}
Here we report the results of our experiments and the ablation studies to assess parameter sensitivity. In particular, we consider the behavior of our model when the \textit{labeled sample per class} (\emph{lpct}) increases.
\subsection{Performance evaluation on VerSe}
We present the accuracy scores for \our on VerSe dataset in Table \ref{tab:VerSe}. We reported our experimental results when one, two and 20 \lpc~ are used. The performances of the intermediate amount of labeled elements (3 to 19) are reported in the ablation study (see Fig.\ref{fig:MFL_GOLD_verse} and Fig.\ref{fig:MFL_PRED_verse}). We highlighted our results in bold when our performances are better than \cite{gella2019disambiguating} and the performances of \cite{gella2019disambiguating} are not in the standard deviation range of our model.

\begin{table*}[ht!]
\centering
\resizebox{\textwidth}{!}{%
\footnotesize
\begin{tabular}{lcclllllll}
\hline
\multicolumn{10}{c}{Using GOLD annotations for objects and captions}          \\ \hline
   &        &         & \multicolumn{3}{c}{Textual} & Visual      & \multicolumn{3}{c}{Concat (CNN+)}                        \\ \cline{4-6} \cline{8-10} 
   & FS$^*$ & MFS$^*$ & O     & C     & O+C   & CNN   & O     & C     & O+C   \\ \hline
Motion - Unsupervised Gella et al. \cite{gella2019disambiguating}    & 60.0   & 76.1    & 35.3  & 53.8  & 55.3  & 58.4  & 48.4  & 66.9  & 58.4  \\
Motion - our (1 lab/sense)            & 60.0   & 76.1    & {\color[HTML]{3531FF} \textbf{62.3$\pm$7.4}}    & 56.6$\pm$8.3    & 58.6$\pm$8.2    & 59.1$\pm$8.3    & {\color[HTML]{3531FF} \textbf{64.7$\pm$5.8}} & 58.8$\pm$7.0    & 59.0$\pm$8.2    \\
Motion - our (2 lab/sense)            & 60.0   & 76.1    & {\color[HTML]{3531FF} \textbf{71.0$\pm$8.8}} & {\color[HTML]{3531FF} \textbf{68.1$\pm$6.3}} & {\color[HTML]{3531FF} \textbf{69.7$\pm$7.9}} & {\color[HTML]{3531FF} \textbf{67.1$\pm$7.6}} & {\color[HTML]{3531FF} \textbf{72.9$\pm$6.9}} & {\color[HTML]{3531FF} \textbf{70.3$\pm$6.1}} & {\color[HTML]{3531FF} \textbf{70.6$\pm$8.1}} \\
Motion - our (16 lab/sense)           & 60.0   & 76.1    & {\color[HTML]{000000} \textbf{90.2$\pm$3.9}} & {\color[HTML]{000000} \textbf{88.5$\pm$0.5}} & {\color[HTML]{000000} \textbf{88.8$\pm$0.1}} & {\color[HTML]{000000} \textbf{90.7$\pm$0.7}} & {\color[HTML]{000000} \textbf{90.0$\pm$3.5}} & {\color[HTML]{000000} \textbf{88.7$\pm$1.3}} & {\color[HTML]{000000} \textbf{88.8$\pm$0.1}} \\
Motion - Supervised Gella et al. \cite{gella2019disambiguating}     & 60.0   & 76.1    & 82.3  & 78.4  & 80.0  & 82.3  & 83.0  & 82.3  & 83.0  \\ \hline
NonMotion - Unsupervised Gella et al. \cite{gella2019disambiguating} & 71.3   & 80.0    & 48.6  & 53.9  & 66.0  & 55.6  & 56.5  & 56.5  & 59.1  \\
NonMotion - our (1 lab/sense)         & 71.3   & 80.0    & 56.6$\pm$13.8   & 54.3$\pm$8.5    & 59.9$\pm$7.1    & 46.3$\pm$9.1    & 57.1$\pm$11.8   & 51.9$\pm$7.1    & 55.6$\pm$8.2    \\
NonMotion - our (2 lab/sense)         & 71.3   & 80.0    & {\color[HTML]{3531FF} \textbf{69.4$\pm$8.9}} & {\color[HTML]{3531FF} \textbf{71.6$\pm$5.7}} & {\color[HTML]{3531FF} \textbf{71.5$\pm$6.7}} & {\color[HTML]{3531FF} \textbf{69.2$\pm$4.3}} & {\color[HTML]{3531FF} \textbf{69.4$\pm$8.8}} & {\color[HTML]{3531FF} \textbf{71.7$\pm$5.5}} & {\color[HTML]{3531FF} \textbf{71.8$\pm$6.8}} \\
NonMotion - our (16 lab/sense)        & 71.3   & 80.0    & {\color[HTML]{000000} \textbf{91.4$\pm$2.4}} & {\color[HTML]{000000} \textbf{90.6$\pm$1.5}} & {\color[HTML]{000000} \textbf{90.2$\pm$1.5}} & {\color[HTML]{000000} \textbf{94.2$\pm$2.9}} & {\color[HTML]{000000} \textbf{91.4$\pm$2.5}} & {\color[HTML]{000000} \textbf{91.0$\pm$2.1}} & {\color[HTML]{000000} \textbf{90.6$\pm$1.5}} \\
NonMotion - Supervised Gella et al. \cite{gella2019disambiguating}   & 71.3   & 80.0    & 79.1  & 79.1  & 79.1  & 80.0  & 80.0  & 80.0  & 80.0  \\ \hline
\hline
\multicolumn{10}{c}{Using PRED annotations for objects and captions}          \\ \hline
   &        &         & \multicolumn{3}{c}{Textual} & Visual      & \multicolumn{3}{c}{Concat (CNN+)}                        \\ \cline{4-6} \cline{8-10} 
   & FS$^*$ & MFS$^*$ & O     & C     & O+C   & CNN   & O     & C     & O+C   \\ \hline
Motion - Unsupervised Gella et al. \cite{gella2019disambiguating}    & 60.0   & 76.1    & 43.8  & 41.5  & 45.3  & 58.4  & 60.0  & 53.0  & 55.3  \\
Motion - our (1 lab/sense)            & 60.0   & 76.1        & {\color[HTML]{3531FF} \textbf{57.3$\pm$5.7}}    & {\color[HTML]{3531FF} \textbf{55.2$\pm$8.1}}    &  {\color[HTML]{3531FF} \textbf{56.1$\pm$7.4}}    & 59.1$\pm$8.1    & 58.2$\pm$6.9    & 58.1$\pm$7.9    & 58.2$\pm$6.3    \\
Motion - our (2 lab/sense)            & 60.0   & 76.1        & {\color[HTML]{3531FF} \textbf{63.0$\pm$9.0}} & {\color[HTML]{3531FF} \textbf{61.2$\pm$7.7}} & {\color[HTML]{3531FF} \textbf{61.4$\pm$8.8}} & {\color[HTML]{3531FF} \textbf{67.1$\pm$7.6}} & 64.9$\pm$8.0    & {\color[HTML]{3531FF} \textbf{62.9$\pm$6.8}} & {\color[HTML]{3531FF} \textbf{64.1$\pm$7.4}} \\
Motion - our (16 lab/sense)           & 60.0   & 76.1        & {\color[HTML]{000000} \textbf{86.9$\pm$4.1}} & {\color[HTML]{000000} \textbf{87.3$\pm$0.2}} & {\color[HTML]{000000} \textbf{87.4$\pm$0.2}} & {\color[HTML]{000000} \textbf{90.6$\pm$0.7}} & {\color[HTML]{000000} \textbf{87.6$\pm$2.4}} & {\color[HTML]{000000} \textbf{87.8$\pm$0.7}} & {\color[HTML]{000000} \textbf{87.6$\pm$0.3}} \\
Motion - Supervised Gella et al. \cite{gella2019disambiguating}                  & 60.0   & 76.1    & 80.0  & 69.2  & 70.7  & 82.3  & 83.0  & 82.3  & 83.0  \\
Motion - Supervised Silberer et al.$^\#$ \cite{silberer2018grounding}              & 71.3   & 80.0   & -  & -  & -  & -  & 84.8 $\pm$ 0.69  & - & -  \\ \hline
NonMotion - Gella et al. \cite{gella2019disambiguating}              & 71.3   & 80.0    & 46.0  & 61.7  & 55.6  & 55.6  & 52.1  & 60.0  & 55.6  \\
NonMotion - our (1 lab/sense)         & 71.3   & 80.0       & 52.4$\pm$10.5   & 55.8$\pm$9.1    & 55.8$\pm$9.0    & 46.3$\pm$9.2    & 53.5$\pm$8.4    & 55.1$\pm$8.1    & 54.9$\pm$7.4    \\
NonMotion - our (2 lab/sense)         & 71.3   & 80.0       & {\color[HTML]{3531FF} \textbf{61.7$\pm$5.6}} & {\color[HTML]{3531FF} \textbf{75.4$\pm$4.2}} & {\color[HTML]{3531FF} \textbf{75.6$\pm$4.2}} & {\color[HTML]{3531FF} \textbf{69.2$\pm$4.3}} & {\color[HTML]{3531FF} \textbf{63.6$\pm$3.4}} & {\color[HTML]{3531FF} \textbf{76.0$\pm$3.4}} & {\color[HTML]{3531FF} \textbf{74.5$\pm$6.2}} \\
NonMotion - our (16 lab/sense)        & 71.3   & 80.0       & {\color[HTML]{000000} \textbf{92.2$\pm$2.8}} & {\color[HTML]{000000} \textbf{93.8$\pm$3.0}} & {\color[HTML]{000000} \textbf{93.0$\pm$3.1}} & {\color[HTML]{000000} \textbf{94.2$\pm$2.9}} & {\color[HTML]{000000} \textbf{92.3$\pm$2.9}} & {\color[HTML]{000000} \textbf{93.8$\pm$3.0}} & {\color[HTML]{000000} \textbf{93.0$\pm$3.1}} \\
NonMotion - Supervised Gella et al. \cite{gella2019disambiguating}              & 71.3   & 80.0   & 78.2  & 77.3  & 77.3  & 80.0  & 80.0  & 80.3  & 80.0  \\
NonMotion - Supervised Silberer et al.$^\#$ \cite{silberer2018grounding}              & 71.3   & 80.0   & -  & -  & -  & -  & 80.4 $\pm$ 0.57  & -  & -  \\ \hline
\end{tabular}}
\caption{
\footnotesize Sense prediction accuracy using \textit{PRED} and \textit{GOLD} settings in VerSe-19verbs for unsupervised, semisupervised and supervised approaches using different types of senses and image representation features.
In \textbf{bold} the results that outperform the supervised method, while in {\color[HTML]{3531FF} \textbf{blue}} the ones outperforming the unsupervised model. $^\#$ uses a different embedding than \cite{gella2019disambiguating}, hence direct comparisons are not straightforward.}
\label{tab:verb19}

\end{table*}

\vspace{-0.5cm}
\paragraph{\our using one labeled sample per class:} As a first step, we investigate the performances of our model considering 1 \lpc. Despite being an extreme case, \our outperformed the unsupervised model of \cite{gella2019disambiguating} on 3 different features over 7 in both motion and non-motion verb classes (see Table \ref{tab:VerSe}). Considering the two heuristics (FS and MFS), \our performed on par with FS while is not able to reach the MFS performances. This confirms the nature of our model, being semi-supervised its performances are typically bounded between unsupervised (FS and \cite{gella2019disambiguating}) and supervised (MFS) methods.
\vspace{-0.5cm}
\paragraph{\our using two labeled samples per class:} 
When adding an extra labeled point, hence having 2 \lpc, \our outperforms the unsupervised state-of-the-art \cite{gella2019disambiguating} and FS heuristic in all features modalities and classes (motion and non-motion) with a large margin. The MFS is still performing better, but considers all the labels.

\vspace{-0.5cm}
\paragraph{\our using 20 labeled sample per class:} In this experiment, we moved to the other extreme in terms of annotated data, providing a lot of labeled samples to our model. The result is that \our significantly outperforms both \cite{gella2019disambiguating}, FS and MFS heuristic in all features types for both motion and non-motion verbs and in both GOLD and PRED settings. This is a remarkable result, in particular because we outperformed MFS, which uses the entire labeled dataset.
\vspace{-0.5cm}
\paragraph{Additional considerations:} It is worth noting that the standard deviation of our experiments, when considering 1 \lpc~ is very high and is getting smaller increasing the labeled set size. This is obvious since we are adding labeled informations to our model. Moreover, although multimodality provides strong performance  gain  in the  PRED  setting  rather than  in the GOLD  setting, it  brings  only a  marginal  improvement  in  \our  compared  to  \cite{gella2019disambiguating}. This might be explained by the nature of the GTG algorithm, which exploits all the possible relations between samples in the datasets, hence the unimodal features might be sufficient. In fact, in general, \our gives better performances considering unimodal features.
\subsection{Performance evaluation on VerSe-19verbs}
\vspace{-0.2cm}
In Table \ref{tab:verb19} we summarized our performances on the VerSe-19verbs for GOLD and PRED settings.
\vspace{-0.5cm}

\paragraph{Using GOLD annotations for objects and captions:} 

As in the VerSe experiment, we tested our model under different \lpc, employing 1, 2 and 16 \lpc.
We started the experiments considering the extreme case in which we have only 1 \lpc. In this case, \our achieves quite low performances, outperforming the unsupervised model of \cite{gella2019disambiguating} only in 2 cases (both considering the O features) and only in the motion class (see Table \ref{tab:verb19}). Speculating, the motivation of these performances might be the following: motion verbs represent typically actions performed between specific entities/objects. For example, consider the verb ”play” associated to an image containing a person and a musical instrument. The association with the correct sense is straightforward due to the two entities. That’s why, in the case of motion verbs, the O’s feature has a strong influence on the overall performances.

The remaining results are comparable (considering the standard deviation) or slightly worst than the competitors. Regarding the heuristics, \our reaches comparable performances to FS only in the motion class, while reaching absolutely unsatisfying results compared to MFS and in the entire non-motion verb class.

When we tested the model with 2 \lpc~ we got something interesting. With just 2 \lpc, we reached better performances than the unsupervised model in both motion and non-motion verb classes. Regarding the heuristics, we achieved comparable or better results than FS, while MFS is still the stronger competitor. It is worth noting that we used only 2 \lpc, hence the annotation effort is dramatically low. As in the VerSe experiment we tested our model on the other extreme case, with 16 \lpc. In this case, we strongly surpass both the heuristics, the unsupervised model and also the supervised ones (where at least 16 \lpc~ are used in training). This is a remarkable results, considering that MFS relies on the entire labeled dataset.

\vspace{-0.2cm}
\paragraph{Using PRED annotations for objects and captions:} We performed the same experiments conducted in the GOLD setting but considering the PRED data. 
Differently from the previous experiment, when 1 \lpc~ is considered, \our clearly outperforms the unsupervised competitor in the motion setting in 3 over 7 cases and performed on par in the remaining. Regarding the non-motion setting, \our is poorly performing and is not able to outperform both the heuristics and the unsupervised model in \cite{gella2019disambiguating}.
As in the GOLD setting, when using just 2 \lpc, the performances start to increase considerably. In this case, \our reaches better performances than the unsupervised model and performed on par or better than the FS heuristics in the motion setting. Considering the non-motion verbs, \our outperformed completely the unsupervised model of \cite{gella2019disambiguating} but is still under-performing with respect to MFS and the supervised model of  \cite{gella2019disambiguating}.

When 16 \lpc~ are used, \our significantly outperforms both supervised \cite{gella2019disambiguating} and the MFS heuristic in both motion and non-motion verbs settings. 

Regarding the PRED setting, another work reported sense prediction performances on VerSe-19verbs. In \cite{silberer2018grounding} the authors used the same logistic classifier and evaluation protocol as in \cite{gella2019disambiguating} but with a different feature embedding, called \emph{ImgObjLoc}. Indeed, \cite{silberer2018grounding} outperformed \cite{gella2019disambiguating} showing that their feature model is more expressive and powerful. Nevertheless, when considering 16 \lpc, our model with standard features outperforms the \cite{silberer2018grounding}. The gap is large, we gain 3 points and more than 10 points in the motion and non motion settings respectively.  We left as future work, applying \our to the features of \cite{silberer2018grounding}.
 
\subsection{Ablation of \our}
\begin{figure*}[th!]%
    \centering
    \subfloat[text]{{\includegraphics[width=.32\textwidth, trim={0.8cm 0.3cm 2cm 1cm},clip]{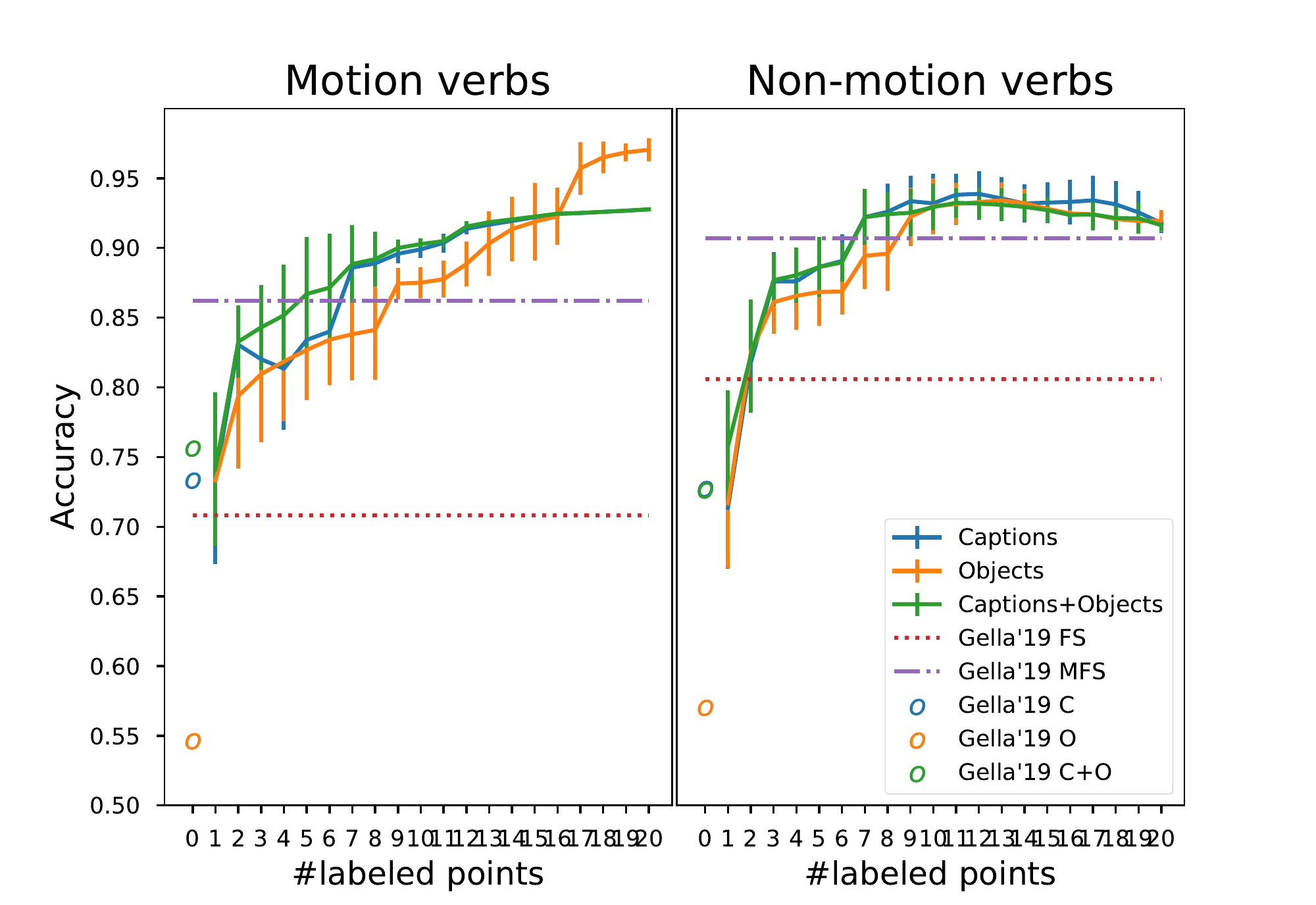} }}%
    \hfill
    \subfloat[cnn]{{\includegraphics[width=.32\textwidth, trim={0.8cm 0.3cm 2cm 1cm},clip]{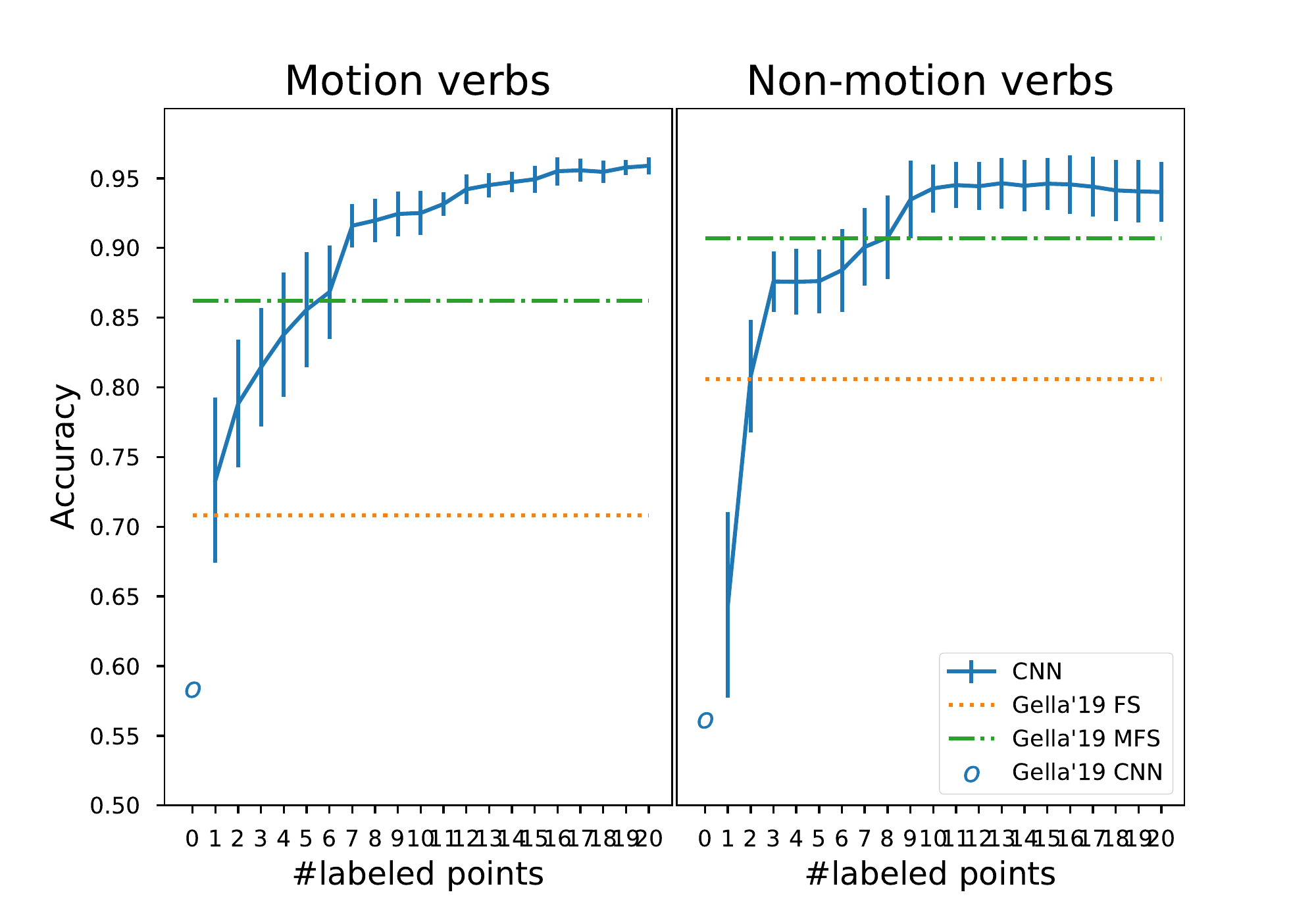} }}%
    \hfill
    \subfloat[cnn+text]{{\includegraphics[width=.32\textwidth, trim={0.8cm 0.3cm 2cm 1cm},clip]{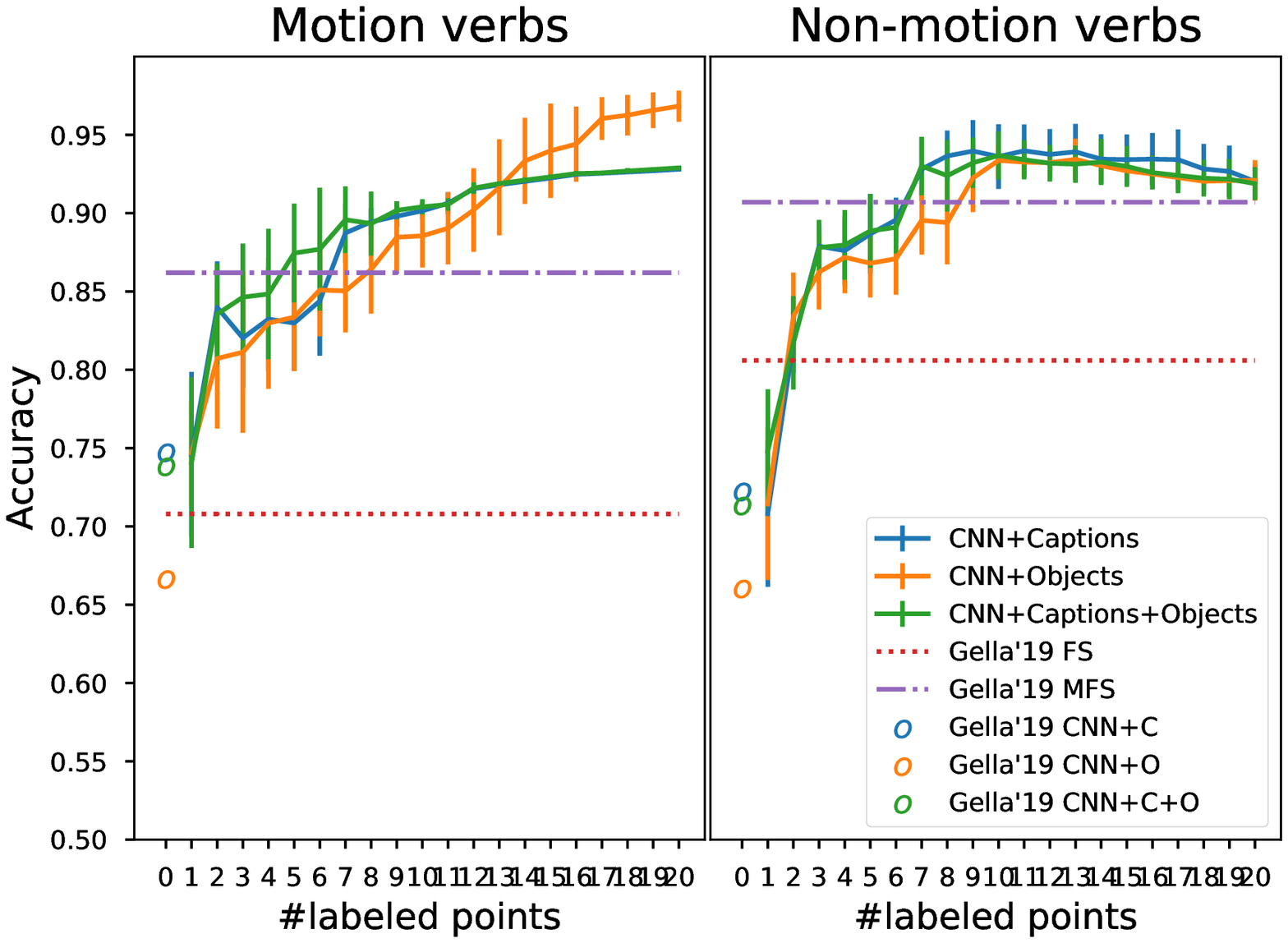} }}
    \vspace{-0.4cm}
    \caption{\footnotesize GOLD results {on VerSe} for text data, cnn and cnn+text varying the number of labeled points in comparison with Gella et al. \cite{gella2019disambiguating} approach (circles), {FS and MFS results from \cite{gella2019disambiguating} (dashed lines)}}%
    \label{fig:MFL_GOLD_verse}%
\end{figure*}

\begin{figure*}[th!]%
    \vspace{-0.4cm}
\centering
    \subfloat[text]{\includegraphics[width=.32\textwidth, trim={0.8cm 0.3cm 2cm 1cm},clip]{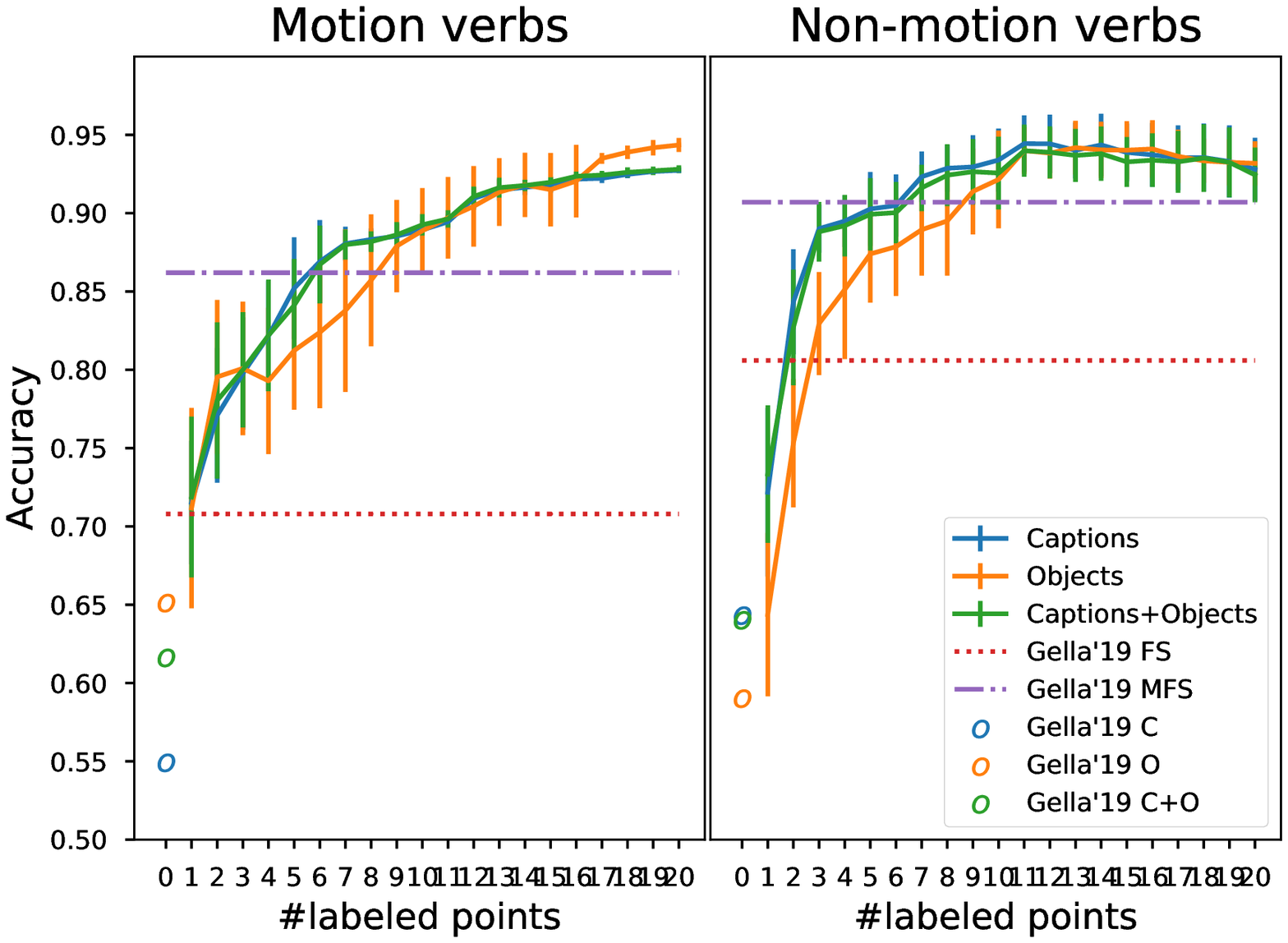} }%
    \hfill
    \subfloat[cnn]{\includegraphics[width=.32\textwidth, trim={0.8cm 0.3cm 2cm 1cm},clip]{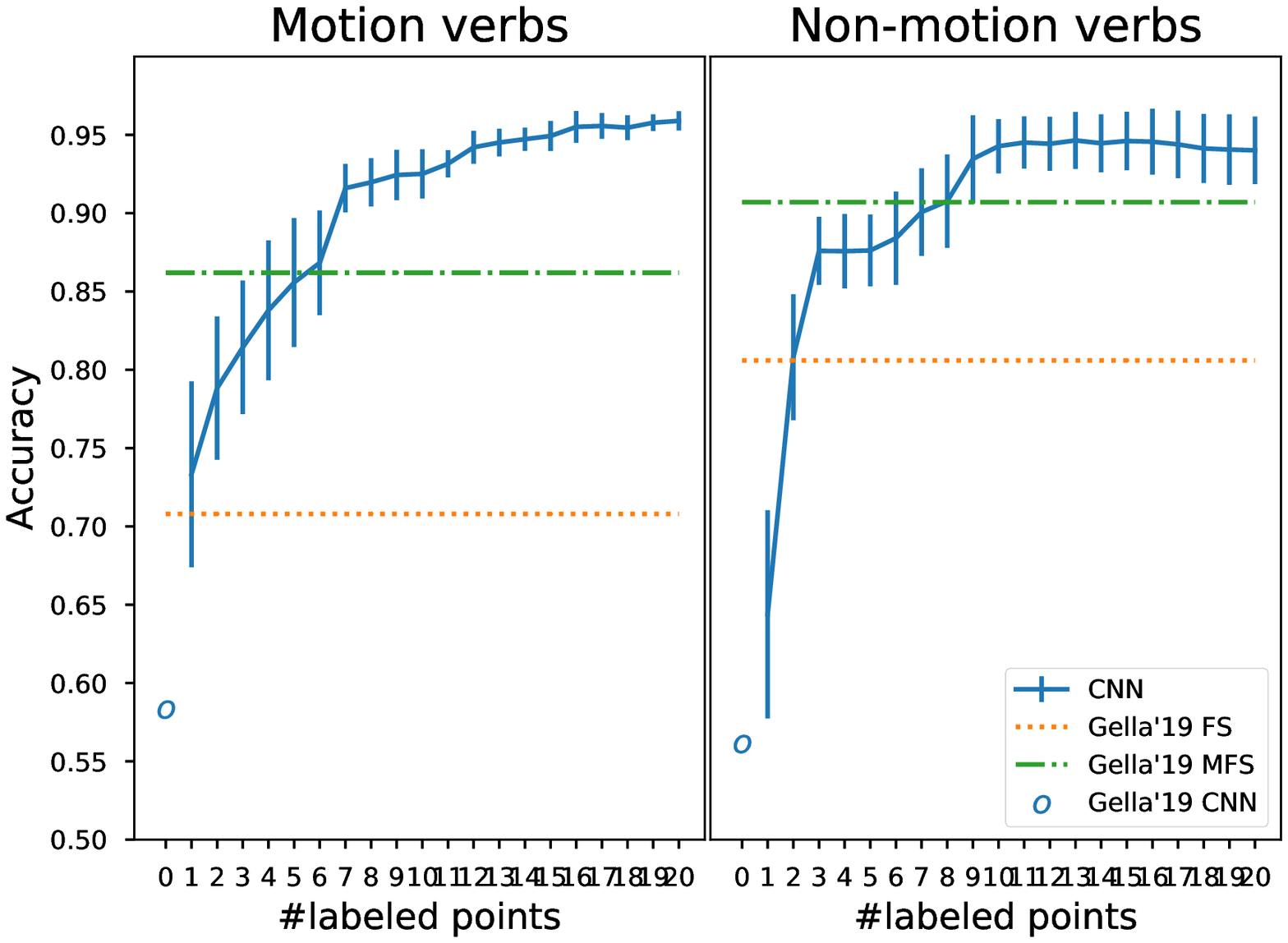} }%
    \hfill   
    \subfloat[cnn+text]{\includegraphics[width=.32\textwidth, trim={0.8cm 0.3cm 2cm 1cm},clip]{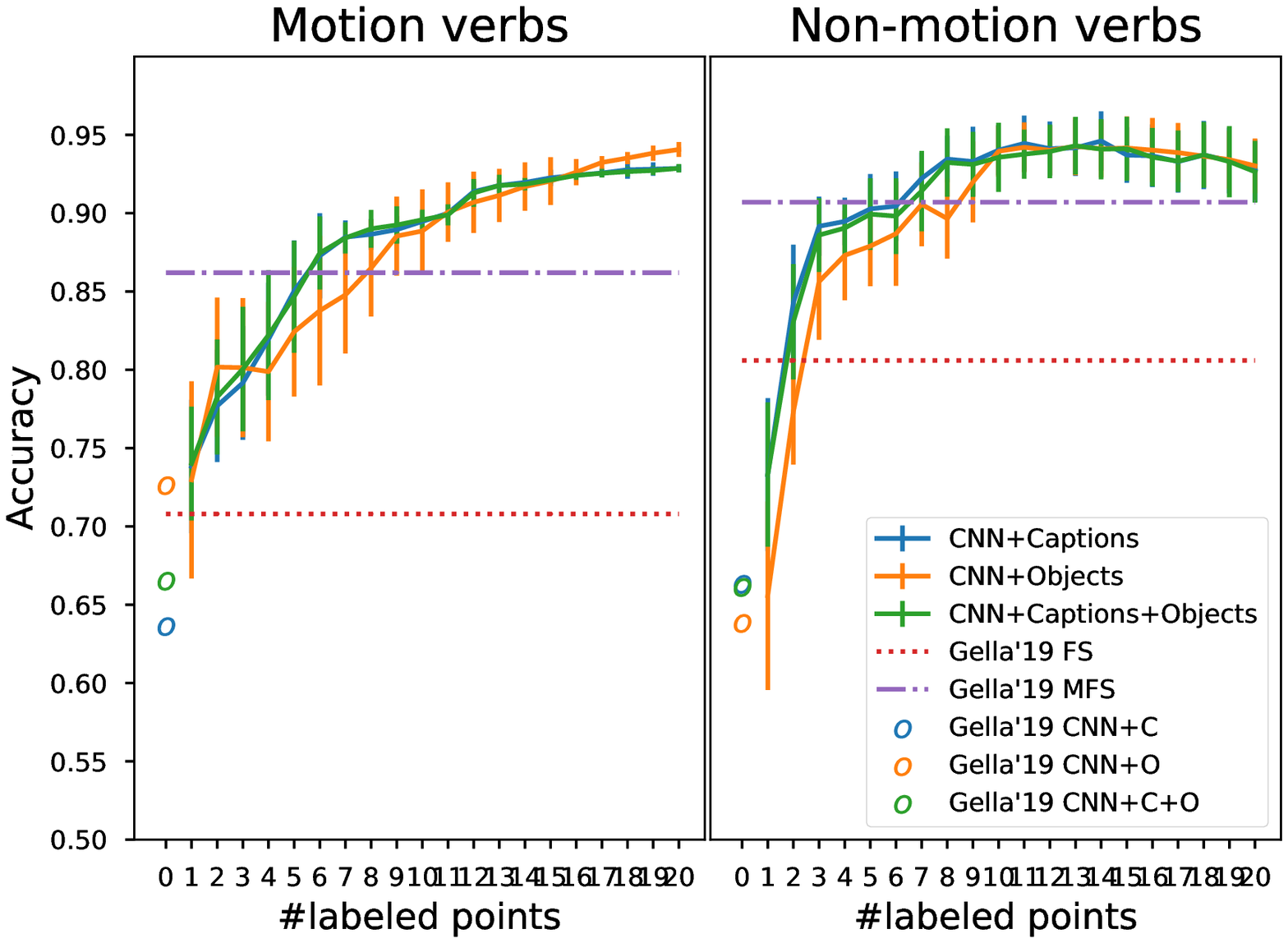} }%
    \vspace{-0.4cm}
    \caption{\footnotesize PRED results {on VerSe} for text data, cnn and cnn+text varying the number of labeled points in comparison with Gella et al. \cite{gella2019disambiguating} approach (circles), {FS and MFS results from \cite{gella2019disambiguating} (dashed lines)}. The central plot is repeated on purpose to avoid a blank figure.}\label{fig:MFL_PRED_verse}
\end{figure*}

\begin{figure*}[th!]%
    \vspace{-0.4cm}
    \centering
    \subfloat[text]{\includegraphics[width=.32\textwidth, trim={0.8cm 0.3cm 2cm 1cm},clip]{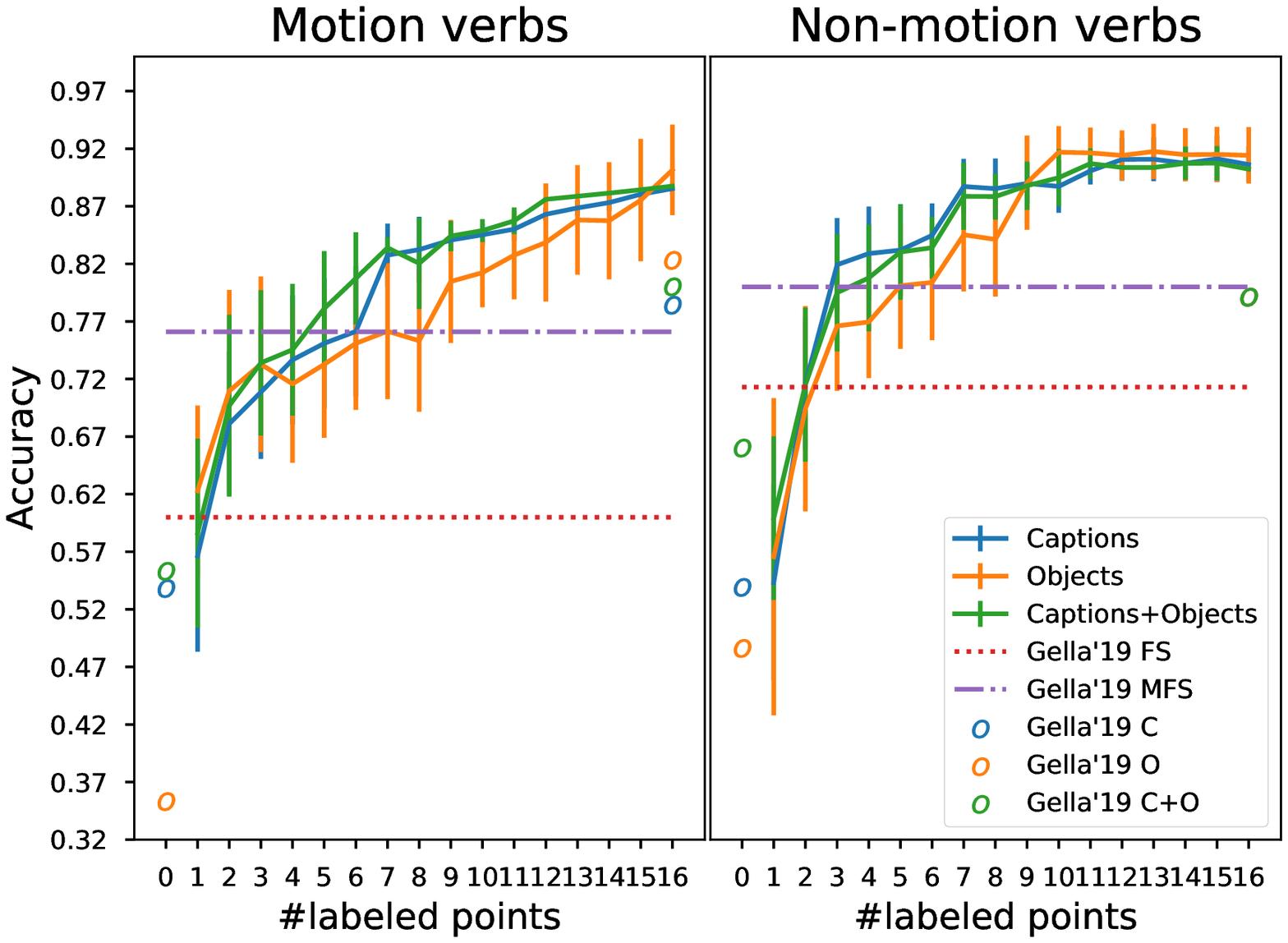}}%
    \hfill
    \subfloat[cnn]{\includegraphics[width=.32\textwidth, trim={0.8cm 0.3cm 2cm 1cm},clip]{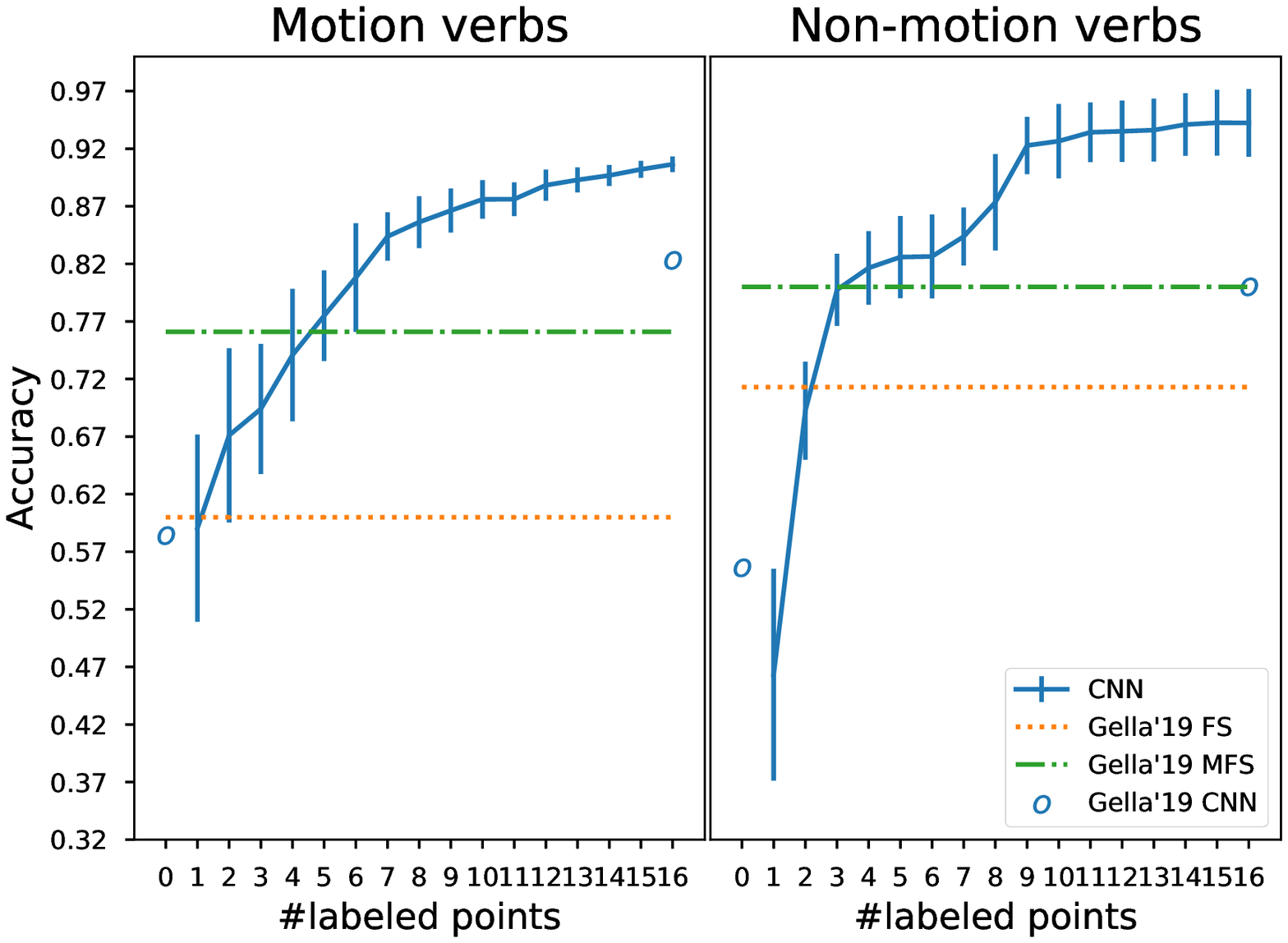} }%
    \hfill
    \subfloat[cnn+text]{\includegraphics[width=.32\textwidth, trim={0.8cm 0.3cm 2cm 1cm},clip]{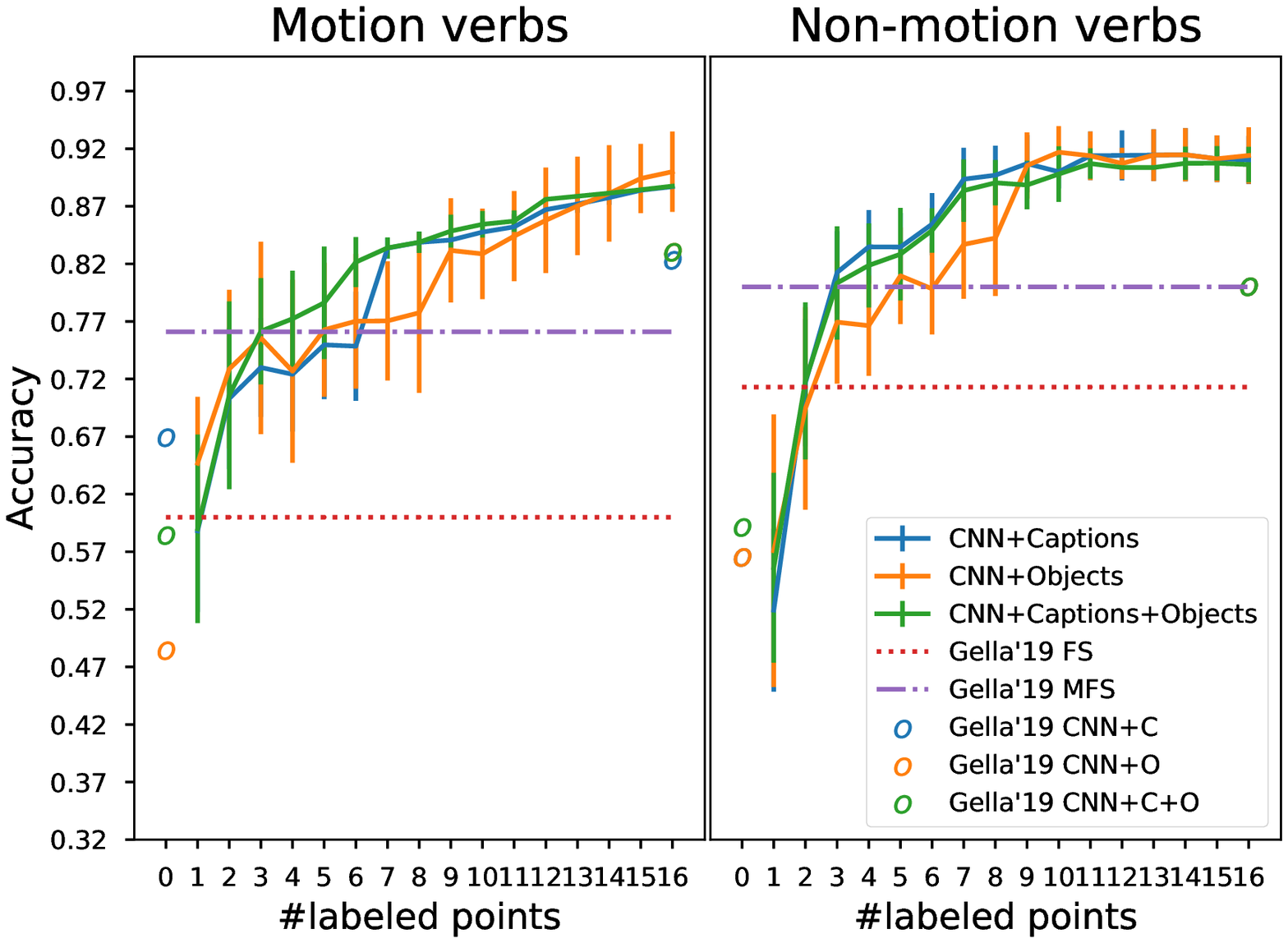} }%
    \vspace{-0.4cm}
    \caption{\footnotesize GOLD results on {VerSe-verb19} for text data, cnn and cnn+text varying the number of labeled points in comparison with Gella et al. \cite{gella2019disambiguating} approach (circles), {FS and MFS results reported from \cite{gella2019disambiguating} (dashed lines).}}%
    \label{fig:MFL_GOLD_verb19}%
\end{figure*}

\begin{figure*}[th!]%
    \vspace{-0.4cm}
    \centering
    \subfloat[text]{\includegraphics[width=.32\textwidth, trim={0.8cm 0.3cm 2cm 1cm},clip]{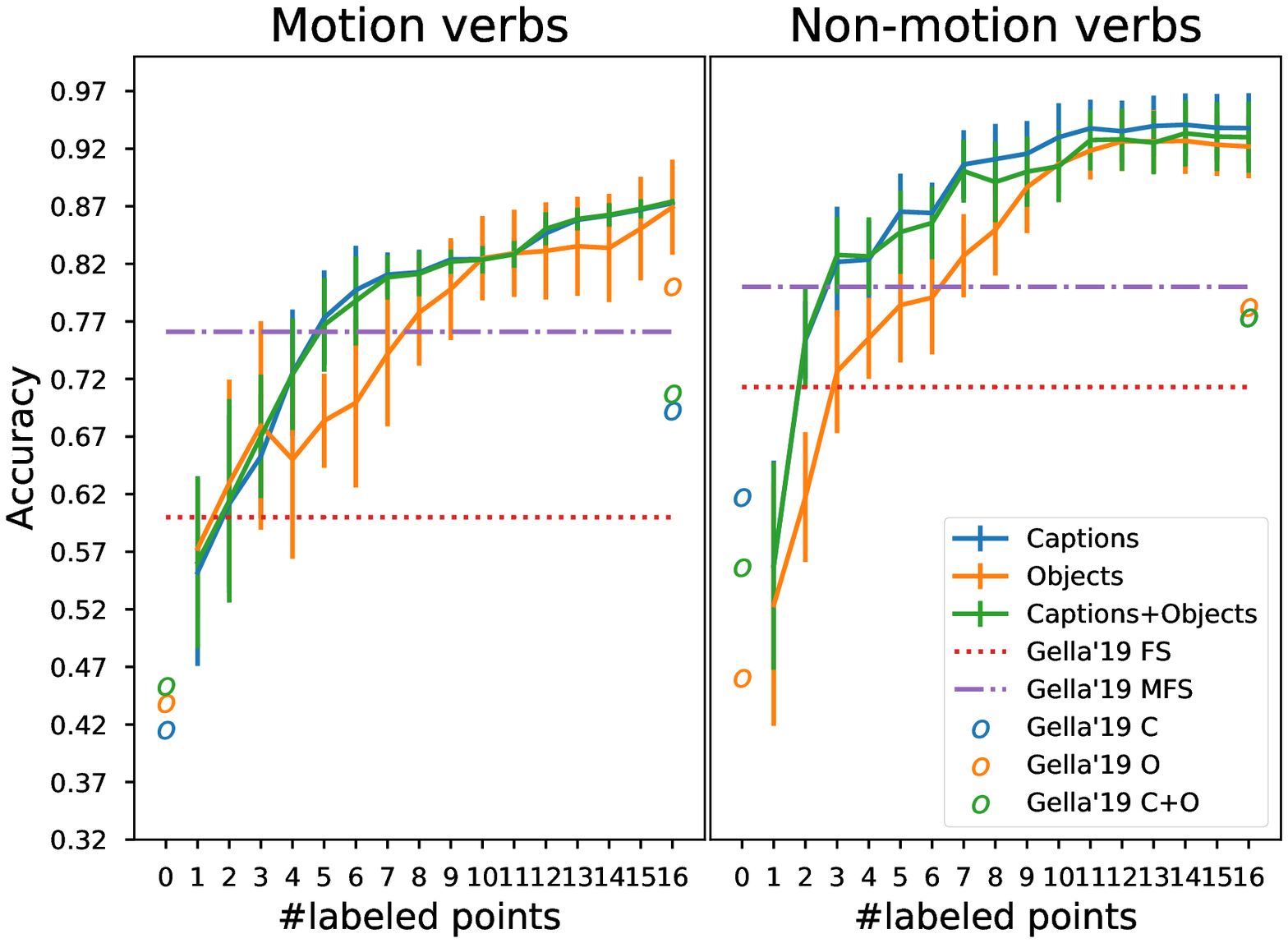}}%
    \hfill
    \subfloat[cnn]{\includegraphics[width=.32\textwidth, trim={0.8cm 0.3cm 2cm 1cm},clip]{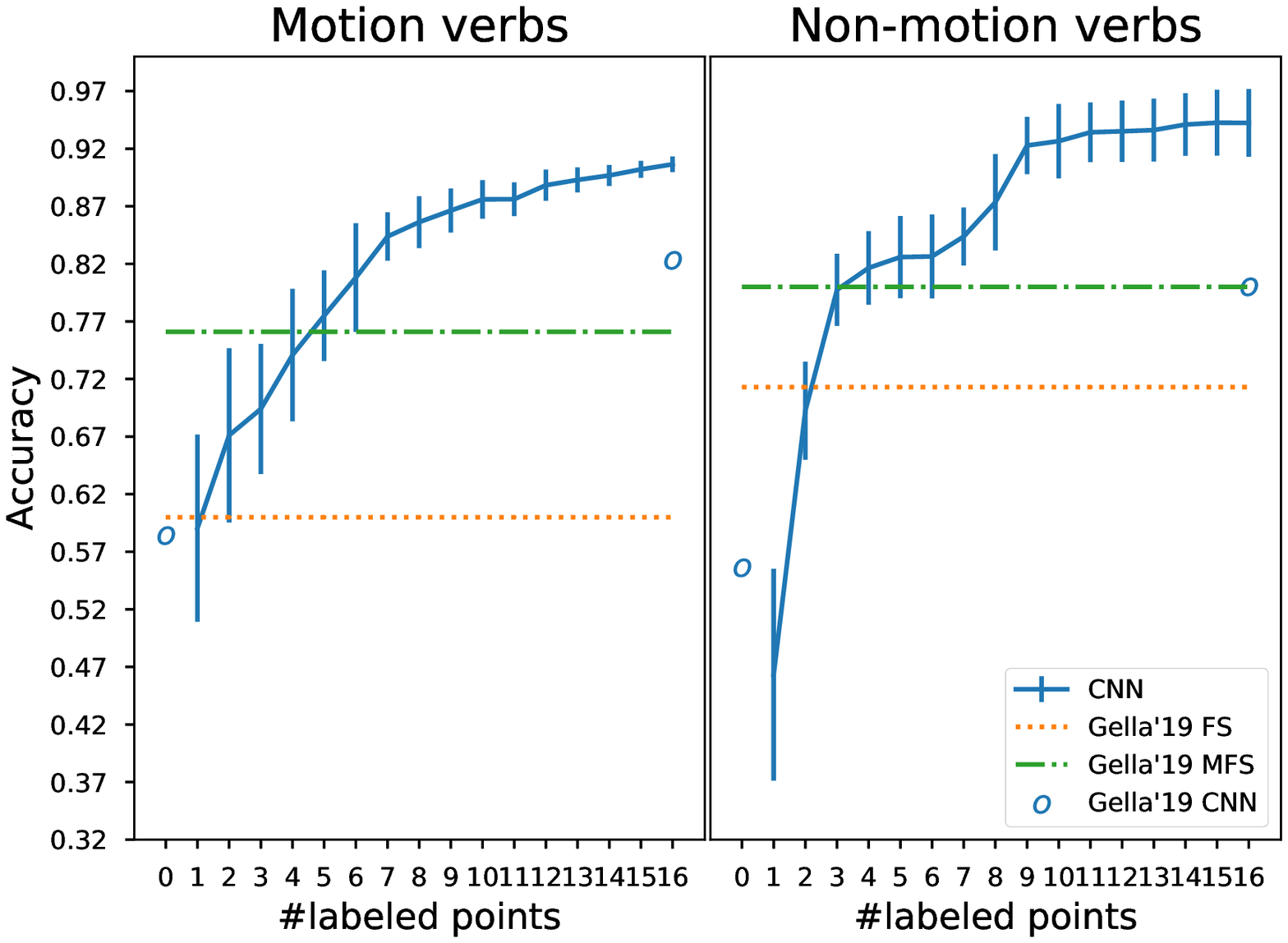}}%
    \hfill
    \subfloat[cnn+text]{\includegraphics[width=.32\textwidth, trim={0.8cm 0.3cm 2cm 1cm},clip]{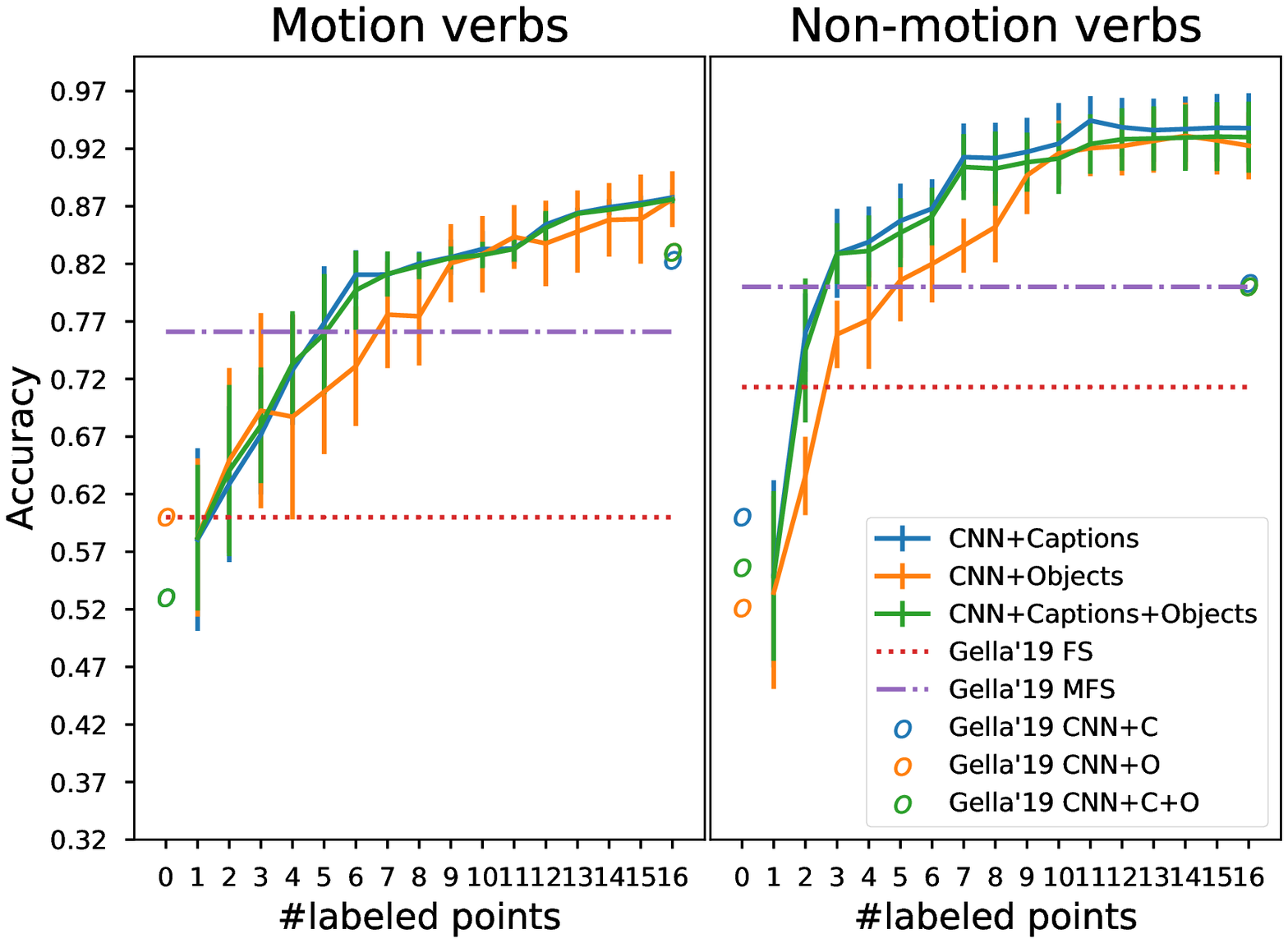}}%
    \vspace{-0.4cm}
    \caption{\footnotesize PRED results {on VerSe-verb19} for text data, cnn and cnn+text varying the number of labeled points in comparison with Gella et al. \cite{gella2019disambiguating} approach (circles), {FS and MFS results reported from \cite{gella2019disambiguating} (dashed lines).}}%
    \label{fig:MFL_PRED_verb19}%
\end{figure*}

\vspace{-0.2cm}
In this ablation study, we reported the performances of \our when the labeled set size increases. This analysis is particularly useful to assess the effort needed for data annotation. The results on the VerSe dataset are shown in Figures \ref{fig:MFL_GOLD_verse} and \ref{fig:MFL_PRED_verse}, while for the VerSe-verb19 are in Figures \ref{fig:MFL_GOLD_verb19} and \ref{fig:MFL_PRED_verb19}. We reported means and standard deviations for all 15 runs. Alongside our results we added the performances of FS, MFS and Unsupervised \cite{gella2019disambiguating} (see $lpc=0$). The supervised results of \cite{gella2019disambiguating} (see $lpc=16$) are reported only for VerSe-verb19.

\vspace{-0.4cm}
\paragraph{Ablation on VerSe dataset}
As can be seen, the performances with 1 labeled point per sense are comparable or better than \cite{gella2019disambiguating} while with 2 or more labeled points we outperform the state-of-the-art \cite{gella2019disambiguating}. This confirms that, for this task, few labeled points are sufficient, hence the labeling effort can be drastically reduced. After around 6 labeled points, we outperfom MFS significantly. 
\SV{In general, we noted that the higher the number of labeled points per sense, the greater the overall accuracy and smaller the standard deviation. The accuracy follows a logarithmic growth, i.e., the variation of the number of labels has a relevant role when they are few, whereas, with more than 6-8 labeled points per class, the accuracy starts converging. For the non-motion verbs, when textual features are used, the performance starts to decrease after reaching to a peak. This shows us that high number of labeled points creates a noise effect at these settings, i.e. similar elements in different classes mislead classification accuracy. 
} We also noted that when visual features are used, the performance converges for the non-motion verbs while it continues to increase for the motion verbs. This was actually expected since actions of motion verbs are apparently more recognizable on images.
\vspace{-0.3cm}
\paragraph{Ablation on VerSe-19verbs dataset}

Both figures \ref{fig:MFL_GOLD_verb19} and \ref{fig:MFL_PRED_verb19} shown similar behaviors to the ablation on the VerSe dataset. We can drawn similar considerations, and noting that after 6-8 \lpc~ the \our model outperforms all the competitors showing a strong stability in the performances. 

\section{Conclusions}
\vspace{-0.2cm}
\SV{In this paper, we proposed a new model for multimodal VVSD tasks based on a transductive semi-supervised learning method. The proposed method is principled, well-funded and outperforms consistently the competitors. The transductive reasoning, used to perform the verb-sense disambiguation, considers the similarity of all the elements in the dataset, reaching a global consensus exploiting a \emph{label consistency} principle. This differs completely from the (small) literature, which still relies on inductive methods that disambiguate visual verb in isolation. The proposed model is general enough to handle both unimodal and multimodal embeddings of $<$image,verb$>$ pairs. Furthermore, we showed that 2 labeled points per sense are sufficient to outperform unsupervised state-of-the-art methods while 6-8 points are enough to obtain better performances than fully-supervised disambiguation models. 
}

\newpage

{\small
\bibliographystyle{ieee_fullname}
\bibliography{egbib}
}

\end{document}